\pdfoutput=1

\documentclass[11pt]{article}

\usepackage[]{acl/acl2023}

\usepackage{times}
\usepackage{latexsym}
\usepackage{graphics}
\usepackage{graphicx}
\usepackage{color}
\usepackage{xcolor}
\usepackage{colortbl}
\usepackage{caption}
\usepackage{amsfonts}
\usepackage{amssymb}
\usepackage{amsmath}
\usepackage{nicefrac}
\usepackage{bbm}
\usepackage{booktabs}
\usepackage{microtype}
\usepackage{multirow}
\usepackage{multicol}
\usepackage{tabu}
\usepackage{amsthm}
\usepackage{footmisc}
\usepackage{soul}
\usepackage{arydshln}
\usepackage{wasysym}
\usepackage{array}
\usepackage{makecell}

\usepackage{algpseudocode}
\usepackage{algorithm}
\usepackage{hyperref}
\usepackage{xspace}
\usepackage{url}
\usepackage{utfsym}
\usepackage{dirtytalk}

\usepackage{enumerate}
\newenvironment{itemize*}%
 {\leftmargini=10pt\begin{itemize}%
  \setlength{\itemsep}{0pt}%
  \setlength{\parskip}{0pt}%
  }%
 {\end{itemize}}
\newenvironment{enumerate*}%
 {\begin{enumerate}%
  \setlength{\itemsep}{0pt}%
  \setlength{\parskip}{0pt}}%
 {\end{enumerate}}

\usepackage{float}
\usepackage{enumitem}
\usepackage{tikz}
\usepackage{tcolorbox}
\usepackage{siunitx}
\usepackage{pifont}
\usepackage{longtable}
\usepackage{supertabular}
\usepackage[fixed]{fontawesome5}

\newtcbox{\hlprimarytab}{on line, rounded corners, box align=base, colback=c3!10,colframe=white,size=fbox,arc=3pt, before upper=\strut, top=-2pt, bottom=-4pt, left=-2pt, right=-2pt, boxrule=0pt}
\newtcbox{\hlsecondarytab}{on line, box align=base, colback=red!10,colframe=white,size=fbox,arc=3pt, before upper=\strut, top=-2pt, bottom=-4pt, left=-2pt, right=-2pt, boxrule=0pt}
\newcommand{\daugshifted}{\raisebox{0.5\depth}{$\uparrow$}}

\newcommand{\daulg}[1]{{\hlsecondarytab{\daugshifted{#1}}}}

\definecolor{my_green}{RGB}{51,102,0}
\definecolor{my_red}{RGB}{204, 0, 0}
\newcommand{\colorcmark}{\textcolor{my_green}{\ding{52}}}
\newcommand{\colorxmark}{\textcolor{my_red}{\ding{55}}}

\newcommand{\ours}{\textsc{AutoAct}}

\usepackage[T1]{fontenc}

\usepackage[utf8]{inputenc}

\usepackage{microtype}

\usepackage{inconsolata}

\usepackage{todonotes}

\newcommand{\revise}[1]{\textcolor{black}{#1}}

\title{\textsc{AutoAct}: Automatic Agent Learning from Scratch for QA via Self-Planning}

\author{
  Shuofei Qiao$^{\spadesuit\heartsuit}$,
  Ningyu Zhang$^{\spadesuit\heartsuit}\footnotemark[1]$,
  Runnan Fang$^{\spadesuit\heartsuit}$,
  Yujie Luo$^{\spadesuit\heartsuit}$,
  \\
  \textbf{Wangchunshu Zhou}$^\clubsuit$,
  \textbf{Yuchen Eleanor Jiang}$^\clubsuit$,
  \textbf{Chengfei Lv}$^\diamondsuit$,
  \textbf{Huajun Chen}$^{\spadesuit\heartsuit}$\thanks{$\quad$Corresponding Author.} \\
  $^\spadesuit$Zhejiang University \\
  $^\heartsuit$Zhejiang University - Ant Group Joint Laboratory of Knowledge Graph \\
  $^\clubsuit$AIWaves Inc. ~
  $^\diamondsuit$Alibaba Group \\
  \fontsize{10.2pt}{0.1\baselineskip}\selectfont \texttt{\{shuofei,zhangningyu\}@zju.edu.cn}
}

\begin{document}
\maketitle
\begin{abstract}

Language agents have achieved considerable performance on various complex question-answering tasks by planning with external tools. Despite the incessant exploration in this field, existing language agent systems still struggle with costly, non-reproducible data reliance and face the challenge of compelling a single model for multiple functions. To this end, we introduce \textbf{\ours}, an automatic agent learning framework for QA that does not rely on large-scale annotated data and synthetic planning trajectories from closed-source models (e.g., GPT-4). Given limited data with a tool library, \textbf{\ours} first automatically synthesizes planning trajectories without any assistance from humans or strong closed-source models. Then, \textbf{\ours} leverages a \textit{division-of-labor} strategy to automatically differentiate based on the target task information and synthesized trajectories, producing a sub-agent group to complete the task. We conduct comprehensive experiments with different LLMs, which demonstrates that \textbf{\ours} yields better or parallel performance compared to various strong baselines. Further analysis demonstrates the effectiveness of the \textit{division-of-labor} strategy, with the trajectory quality generated by {\ours} generally outperforming that of others\footnote{Code: \url{https://github.com/zjunlp/AutoAct}.}.

\end{abstract}

\section{Introduction}

Language agents \cite{renda/agent/survey,fudan/agent/survey,multi/agent/survey}, which leverage the powerful reasoning capabilities \cite{shuofei/survey,shangjiao/agent/survey} of Large Language Models (LLMs) to interact with executable tools, have emerged as essential components of AI systems designed to address complex question-answering tasks \cite{autogpt,gpt-engineer,babyagi,medagent,openagents}.
The process of endowing LLMs with such interactive capabilities is referred to as \textit{Agent Learning} wherein \textit{planning} \cite{planning/survey} plays a pivotal role, which is responsible for decomposing complex questions into simpler ones \cite{cot,react,xagent,chatdev}, invoking external tools \cite{hugginggpt,chameleon,toolllm}, reflecting on past mistakes \cite{reflexion,self-refine}, and aggregating information from various sources to reach the final answer.
There have been a lot of works \cite{camel,hugginggpt,metagpt,igas,agentverse,autoagents} that directly prompt closed-source off-the-shelf LLMs to plan on particular tasks.
Despite their convenience and flexibility, closed-source LLMs inevitably suffer from unresolved issues, as their accessibility often comes at a steep price and their black-box nature makes the result reproduction difficult.
In light of this, some recent endeavors have shifted their focus towards imbuing open-source models with planning capabilities through fine-tuning~\cite{fireact,agenttuning,lumos}.

\begin{figure}[t!]
    \centering
    \resizebox{.48\textwidth}{!}{
    \includegraphics{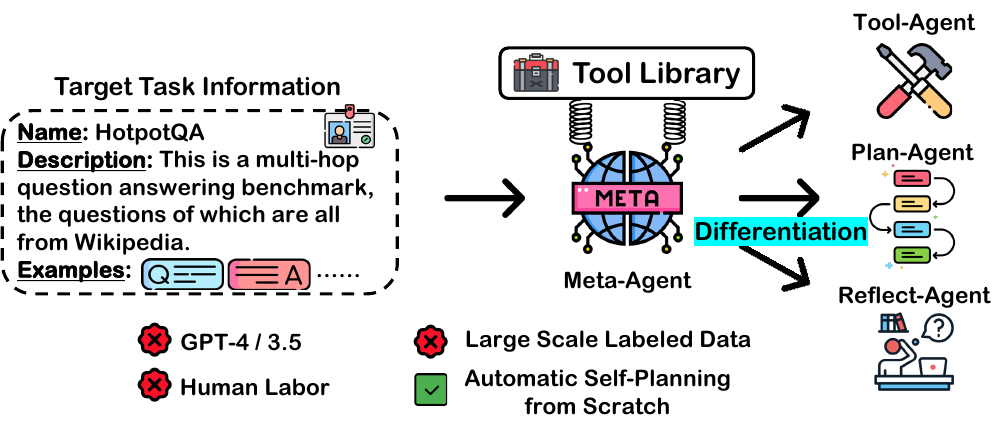}}
    \caption{
    \textbf{The basic framework of {\ours}.}
    Armed with just one tool library, the \textsc{Meta-Agent} can automatically differentiate based on the target task information and produce a sub-agent group that can collaborate to complete the task.
    }
    \label{fig:first}
\end{figure}

However, despite the achievements of the existing fine-tuning-based methods, they are not without limitations.
\textbf{On the one hand}, training open-source models necessitates a substantial amount of annotated QA data pairs and still relies on closed-source models to synthesize planning trajectories.
However, fulfilling these requirements in many real-world scenarios, such as private personal bots or sensitive company business, often proves to be rocky.
\textbf{On the other hand}, from the perspective of agent framework, fine-tuning-based methods compel one single language agent to learn all planning abilities, placing even greater pressure on them.
These contradict Simon's principle of bounded rationality \citep{simon}, which states that \textit{``precise social \textbf{division-of-labor} and clear individual tasks can compensate for the limited ability of individuals to process and utilize information''}.

To this end, we introduce \textbf{\ours}, an automatic agent learning framework for QA, which does not rely on large-scale annotated data and synthetic trajectories from closed-source models while incorporating explicit individual tasks with precise \textit{division-of-labor} (see Fig.~\ref{fig:first}).
Given a limited set of user-provided data examples, {\ours} starts with a \textsc{Meta-Agent} to obtain an augmented database through self-instruct \citep{self-instruct}.
Then, armed with a prepared tool library, the \textsc{Meta-Agent} can automatically synthesize planning trajectories without any assistance from humans or strong closed-source models.
Finally, we propose the \textit{division-of-labor} strategy which resembles \textit{cell differentiation} based on the self-synthesized trajectories (\textit{genes}), where the \textsc{Meta-Agent} acts as a \textit{stem cell} \citep{cell} and differentiates into three sub-agents with distinct functions: task decomposition, tool invocation, and self-reflection, respectively.
Our differentiation process is essentially a parameter-efficient training process on the self-synthesized trajectories with low-consumption resources.
We list the differences between {\ours} and prior works in Tab.~\ref{tab:compare}.

Experiments on complex question-answering tasks with different LLMs demonstrate that {\ours} yields better or parallel performance compared to various strong baselines.
Extensive empirical analysis demonstrates the effectiveness of our appropriate \textit{division-of-labor} strategy.

\section{\ours}

\begin{figure*}[t!]
    \centering
    \resizebox{0.9\textwidth}{!}{
    \includegraphics{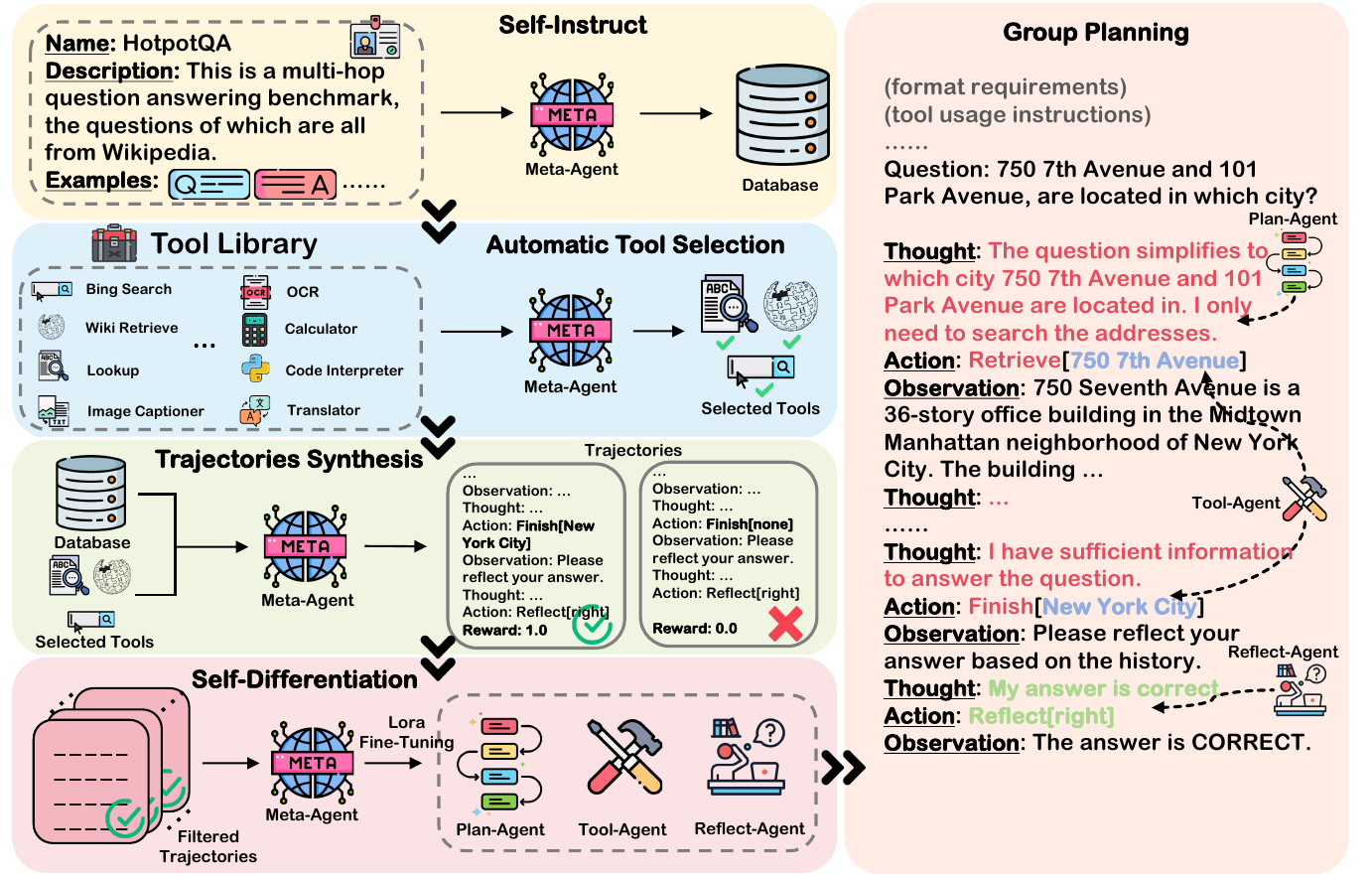}}
    \caption{
    \textbf{The overview of our proposed framework {\ours}.}
    We initiate with \textbf{self-instruct} to extend the task database from scratch.
    Then \textbf{self-planning} is applied to conduct automatic agent learning, including \textit{automatic tool selection}, \textit{trajectories synthesis}, \textit{self-differentiation} and \textit{group planning}.
    Our self-differentiation is a parameter-efficient fine-tuning process to achieve resource-efficient learning.
    }
    \label{fig:method}
\end{figure*}


\subsection{Critical Components of \ours}

\paragraph{\textsc{Meta-Agent}.}
The \textsc{Meta-Agent} is responsible for all the preparatory work before self-differentiation and serves as the backbone model for all sub-agents.
Given limited target task information and a pre-prepared tool library, the \textsc{Meta-Agent} can differentiate into an agent group capable of collaborating to accomplish the target task.
In {\ours}, the \textsc{Meta-Agent} can be initialized with any kind of open-source model.

\paragraph{Target Task Information.}
In this paper, we mainly focus on agent learning from scratch, which means the task information at hand is quite limited, primarily encompassing three aspects:
task name $\mathcal{M}$,
task description $\mathcal{P}$,
task data examples $\mathcal{C}$.
Concretely, $\mathcal{P}$ represents a detailed description of the task's characteristics.
$\mathcal{C}=\{q_i,a_i\}_{i=1}^{|\mathcal{C}|}$ indicates $|\mathcal{C}|$ question-answer example pairs of the task, where $|\mathcal{C}|$ is very small which users can effortlessly provide (e.g., a few demonstrations).
For a more in-depth view of task information, please refer to Appx.~\ref{app:task}.
Note that the task information serves as the only user-provided knowledge of the task for {\ours} to conduct automatic agent learning.

\paragraph{Tool Library.}
To facilitate our agents in automatic task planning, we provide a comprehensive tool library at their disposal.
The tool library can be denoted as $\mathcal{T}=\{m_i,d_i,u_i\}_{i=1}^{|\mathcal{T}|}$, where $m$ represents the tool name, $d$ defines the tool functionality, $u$ details the tool usage instruction, and $|\mathcal{T}|$ stands for the tool amount of the library.
In our automatic procedure, the \textsc{Meta-Agent} has the autonomy to select appropriate tools from the tool library based on the task information.
Users also have the option to expand the tool library according to their specific needs, allowing for more flexible utilization.
We list the details of our tool library in Appx.~\ref{app:tool}.

\subsection{Starting from Scratch via Self-Instruct}
To acquire a sufficient amount of task data and provide an ample training resource, it is necessary to augment the data based on the examples at hand.
We accomplish this process through self-instruct.
Initially, the database $\mathcal{D}$ is set to be equal to the task data examples $\mathcal{C}$, with $\mathcal{C}$ as the seed for data generation.
In each round, the \textsc{Meta-Agent} generates new question-answer pairs by few-shot prompting, and the few-shot prompt examples are randomly sampled from $\mathcal{D}$.
The generated data will be added to $\mathcal{D}$ followed by filtering, with the exclusion of format erroneous and duplicate data before its inclusion.
Eventually, we obtain a database $\mathcal{D}=\{q_i,a_i\}_{i=1}^{|\mathcal{D}|}$, where the number of data $|\mathcal{D}|$ satisfies $|\mathcal{D}|\gg|\mathcal{C}|$.
The prompt we use for self-instruct can be seen in Appx.~\ref{app:prompt_self_instruct} and we list some cases generated through self-instruct in Appx.~\ref{app:database}.

\subsection{Automatic Agent Learning via Self-Planning}

\paragraph{Automatic Tool Selection.}
With the tool library at hand, we ask the \textsc{Meta-Agent} to select applicable tools for each task automatically.
Specifically, we put $\mathcal{T}=\{m_i,d_i,u_i\}_{i=1}^{|\mathcal{T}|}$ in the form of a tool list as part of the prompt.
Along with $\mathcal{T}$, the prompt also includes the task's description $\mathcal{C}$.
Finally, we instruct the \textsc{Meta-Agent} to select an appropriate set of tools $\mathcal{T}_s$ ($\mathcal{T}_s \subset \mathcal{T}$) to wait for synthesizing trajectories.
The prompt we use for automatic tool selection can be seen in Appx.~\ref{app:prompt_tool}.

\paragraph{Trajectories Synthesis.}
Without depending on closed-source models, we enable the \textsc{Meta-Agent} to synthesize planning trajectories on its own.
Equipped with $\mathcal{T}_s$, we instruct the \textsc{Meta-Agent} to synthesize trajectories in a zero-shot manner on the database $\mathcal{D}$ adhering to the format of \texttt{Thought-Action-Observation} as defined in \citet{react}.
In order to obtain high-quality synthesized trajectories, we filter out all the trajectories with $\texttt{reward}<1$ and collect trajectories with exactly correct answers ($\texttt{reward}=1$) as the training source for self-differentiation.
The prompt for trajectories synthesis can be seen in Appx.~\ref{app:prompt_traj_syn}.

\paragraph{Self-Differentiation.}
In order to establish a clear \textit{division-of-labor}, we leverage synthesized planning trajectories to differentiate the \textsc{Meta-Agent} into three sub-agents with distinct functionalities:
\begin{itemize*}
    \item {\small \faTasks} \textbf{\textsc{Plan-Agent}} $\pi_{\rm plan}$ undertakes question decomposition and determines which tool to invoke in each planning loop (Eq.~\ref{eq:plan}).
    \item {\small \faTools} \textbf{\textsc{Tool-Agent}} $\pi_{\rm tool}$ is responsible for how to invoke the tool (Eq.~\ref{eq:tool}) by deciding the parameters for the tool invocation.
    \item {\small \faUserCheck} \textbf{\textsc{Reflect-Agent}} $\pi_{\rm reflect}$ engages in reflection by considering all the historical trajectories and providing a reflection result (Eq.~\ref{eq:reflect}).
\end{itemize*}
We assume that the planning loop at time $t$ can be denoted as $(\tau_t, \alpha_t, o_t)$, where $\tau$ denotes \texttt{Thought}, $\alpha$ signifies \texttt{Action}, and $o$ represents \texttt{Observation}.
$\alpha$ can be further expressed as $(\alpha^m, \alpha^p)$, where $\alpha^m$ is the name of the action, and $\alpha^p$ is the parameters required to perform the action.
Then the historical trajectory at time $t$ can be signaled as:
\begin{align}
\mathcal{H}_t=(\tau_0,\alpha_0,o_0,\tau_1,...,\tau_{t-1},\alpha_{t-1},o_{t-1}).
\end{align}
Eventually, supposing that the prompts of target task information, planning format requirements, and the question are all combined as $\mathcal{S}$, the responsibilities of each sub-agent can be defined as:
\begin{align}
    \label{eq:plan}
    \tau_t, \alpha_t^m &= \pi_{\rm plan}(\mathcal{S}, \mathcal{T}_s, \mathcal{H}_t), \\
    \label{eq:tool}
    \alpha_t^p &= \pi_{\rm tool}(\mathcal{S}, \mathcal{T}_s, \mathcal{H}_t, \tau_t, \alpha_t^m), \\
    \label{eq:reflect}
    \tau^r, \alpha^r &= \pi_{\rm reflect}(\mathcal{S}, \mathcal{T}_s, \mathcal{H}),
\end{align}
where $\tau^r$ and $\alpha^r$ represent the thought and action of the reflection process, and $\mathcal{H}$ is the planning history after finishing the answer.
The trajectories can be reorganized based on the responsibilities above and fed to the \textsc{Meta-Agent} for self-differentiation.
Our differentiation is a parameter-efficient fine-tuning process to achieve resource-efficient learning.
\revise{We give examples of the training data for each sub-agent in Appx.~\ref{app:training_data_example}.}
Particularly, for each sub-agent, we train a specific LoRA \cite{lora}.

\paragraph{Group Planning.}
At inference time, once the tool name $\alpha_t^m$ generated by the \textsc{Plan-Agent} is triggered at time $t$, the \textsc{Tool-Agent} is roused to decide the parameters $\alpha_t^p$ transferred to the specific tool.
The return result of the tool is treated as the observation $o_t$ and handed to the \textsc{Plan-Agent}.
After the collaboration between the \textsc{Plan-Agent} and \textsc{Tool-Agent} reaches a prediction, the \textsc{Reflect-Agent} comes to reflect on the history and provide a reflection result contained in the reflection action $\alpha^r$.
If the reflection result indicates that the prediction is correct, the whole planning process ends.
Otherwise, the \textsc{Plan-Agent} and \textsc{Tool-Agent} will continue the planning based on the reflection information.
The specific sequence of the group planning process can be found in the example on the right of Fig.~\ref{fig:method}.

\begin{table*}[t!]
\centering
\renewcommand\arraystretch{1.0}
\scalebox{0.86}{
\begin{tabular}{clcccccccc}
\hline
\toprule
{\multirow{2}{*}{\textbf{Backbone}}}
& {\multirow{2}{*}{\textbf{Method}}} 
& \multicolumn{4}{c}{\textbf{HotpotQA}} 
& \multicolumn{4}{c}{\textbf{ScienceQA}} \\
\cmidrule(lr){3-6} \cmidrule(lr){7-10}
& & \textbf{Easy} & \textbf{Medium} & \textbf{Hard} & \textbf{All} & \textbf{G1-4} & \textbf{G5-8} & \textbf{G9-12} & \textbf{All} \\
\Xhline{1px}
\multirow{2}{*}{\makecell{GPT-3.5\\Turbo}} & {\small \faToggleOff} {\small \faUser} CoT & 48.21 & 44.52 & 34.22 & 42.32 & 60.83 & 55.83 & 65.00 & 60.56\\
& {\small \faToggleOff} {\small \faUser} Zero-Shot Plan* & 50.71 & 45.17 & 38.23 & 44.70
& 76.67 & 61.67 & 78.33 & 72.22 \\
\Xhline{1px}

\multirow{7}{*}{\makecell{Mistral-7B\\Instruct-v0.2}}
& {\small \faToggleOff} {\small \faUser} CoT & 33.70 & 22.38 & 22.14 & 26.07
& 54.17 & 50.00 & 60.00 & 54.72 \\
& {\small \faToggleOff} {\small \faUser} ReAct & 38.09 & 27.57 & 22.05 & 29.24
& 63.33 & 58.33 & 62.50 & 61.39 \\
& {\small \faToggleOff} {\small \faUser} Chameleon & 37.07 & 26.67 & 19.20 & 27.65
& 65.83 & 62.50 & 66.67 & 65.00 \\
& {\small \faToggleOff} {\small \faUser} Reflexion & 40.78 & \underline{35.02} & 28.36 & 34.72
& \underline{67.50} & \underline{65.83} & \underline{69.17} & \underline{67.50} \\
& {\small \faToggleOff} {\small \faUsers} BOLAA & 40.86 & 32.11 & 22.36 & 31.78
& 64.17 & 61.67 & 65.83 & 63.89 \\
& {\small \faToggleOff} {\small \faUsers} ReWOO & 38.42 & 31.89 & 25.98 & 32.10
& 60.83 & 58.33 & 64.17 & 61.11 \\
& {\small \faToggleOn} {\small \faUser} FireAct & \underline{45.52} & 32.02 & \underline{30.17} & \underline{35.90}
& 65.00 & 62.50 & 64.17 & 63.89 \\
& {\small \faToggleOn} {\small \faUsers} \textbf{\ours} & \textbf{48.69} & \textbf{36.65} & \textbf{31.37} & \textbf{38.89}
& \textbf{69.17} & \textbf{68.33} & \textbf{72.50} & \textbf{70.00}  \\
\Xhline{1px}
\multirow{7}{*}{\makecell{Llama-2\\13B-chat}}
& {\small \faToggleOff} {\small \faUser} CoT & 37.90 & 25.28 & 21.64 & 28.27
& 61.67 & 52.50 & 69.17 & 61.11 \\
& {\small \faToggleOff} {\small \faUser} ReAct & 28.68 & 22.15 & 21.69 & 24.17
& 57.50 & 51.67 & 65.00 & 58.06 \\
& {\small \faToggleOff} {\small \faUser} Chameleon & 40.01 & 25.39 & 22.82 & 29.41
& \underline{69.17} & 60.83 & \underline{73.33} & 67.78 \\
& {\small \faToggleOff} {\small \faUser} Reflexion & 44.43 & 37.50 & \underline{28.17} & 36.70 & 67.50 & \underline{64.17} & 73.33 & \underline{68.33} \\
& {\small \faToggleOff} {\small \faUsers} BOLAA & 33.23 & 25.46 & 25.23 & 27.97
& 60.00 & 54.17 & 65.83 & 60.00 \\
& {\small \faToggleOff} {\small \faUsers} ReWOO & 30.09 & 24.01 & 21.13 & 25.08 & 57.50 & 54.17 & 65.83 & 59.17 \\
& {\small \faToggleOn} {\small \faUser} FireAct & \underline{45.83} & \underline{38.94} & 26.06 & \underline{36.94}
& 60.83 & 57.50 & 67.50 & 61.94 \\
& {\small \faToggleOn} {\small \faUsers} \textbf{\ours} & \textbf{47.29} & \textbf{41.27} & \textbf{32.92} & \textbf{40.49}
& \textbf{70.83} & \textbf{66.67} & \textbf{76.67} & \textbf{71.39} \\
\Xhline{1px}
\multirow{7}{*}{\makecell{Llama-2\\70B-chat}}
& {\small \faToggleOff} {\small \faUser} CoT & 45.37 & 36.33 & 32.27 & 37.99 &
74.17 & 64.17 & 75.83 & 71.39 \\
& {\small \faToggleOff} {\small \faUser} ReAct & 39.70 & 37.19 & 33.62 & 36.83
& 64.17 & 60.00 & 72.50 & 65.56 \\
& {\small \faToggleOff} {\small \faUser} Chameleon & 46.86 & 38.79 & 34.43 & 40.03
& \underline{77.83} & \underline{69.17} & 76.67 & \underline{74.56} \\
& {\small \faToggleOff} {\small \faUser} Reflexion & 48.01 & \underline{46.35} & 35.64 & \underline{43.33} & 75.83 & 67.50 & \underline{78.33} & 73.89 \\
& {\small \faToggleOff} {\small \faUsers} BOLAA & 46.44 & 37.29 & 33.49 & 39.07
& 70.00 & 67.50 & 75.00 & 70.83 \\
& {\small \faToggleOff} {\small \faUsers} ReWOO & 42.00 & 39.58 & 35.32 & 38.96 & 65.00 & 61.67 & 76.67 & 67.78 \\
& {\small \faToggleOn} {\small \faUser} FireAct & \underline{50.82} & 41.43 & \underline{35.86} & 42.70
& 72.50 & 68.33 & 75.00 & 71.94 \\
& {\small \faToggleOn} {\small \faUsers} \textbf{\ours} & \textbf{56.94} & \textbf{50.12} & \textbf{38.35} & \textbf{48.47}
& \textbf{82.50} & \textbf{72.50} & \textbf{80.83} & \textbf{78.61}  \\
\bottomrule
\hline
\end{tabular}
}
\caption{
\textbf{Main results of {\ours} compared to various baselines} on HotpotQA and ScienceQA.
The icon {\small \faToggleOff} indicates prompt-based agent learning without fine-tuning, while {\small \faToggleOn} means fine-tuning-based agent learning.
{\small \faUser} denotes single-agent learning and {\small \faUsers} symbolizes multi-agent learning.
The best results of each model are marked in \textbf{bold} and the second-best results are marked with \underline{underline}.
*We compare the zero-shot plan performance of GPT-3.5-Turbo to ensure fairness in our evaluation since our setup does not include annotated trajectory examples.
}
\label{tab:main_results}
\end{table*}

\section{Experimental Setup}

\paragraph{Tasks and Metrics.}
\label{para:task}
We evaluate {\ours} on HotpotQA \cite{hotpotqa} and ScienceQA \cite{scienceqa}.
HotpotQA is a multi-hop QA task challenging for rich background knowledge, the answer of which is usually a short entity or yes/no.
Following \citet{bolaa}, we randomly select 300 dev questions divided into three levels for evaluation, with 100 questions in each level.
For HotpotQA, the $\texttt{reward} \in [0,1]$ is defined as the F1 score grading between the prediction and ground-truth answer.
ScienceQA is a multi-modal QA task spanning various scientific topics.
We also divide the test set into three levels based on the grade, with 120 randomly sampled data in each level.
Since ScienceQA is a multi-choice task, the $\texttt{reward} \in \{0,1\}$ is exactly the accuracy.
Note that due to the limitations of LMs in generating images, for ScienceQA, during the self-instruct stage, we directly generate captions for the images instead.

\paragraph{Baselines.}
We choose the open-source Llama-2 models \cite{llama2} \revise{and Mistral-7B \cite{mistral}} as the backbones of our \textsc{Meta-Agent} and sub-agents.
The compared baselines include \textbf{CoT} \cite{cot}, \textbf{\textsc{ReAct}}, \textbf{Chameleon} \cite{chameleon}, \textbf{Reflexion} \cite{reflexion}, \textbf{BOLAA} \cite{bolaa}, \textbf{ReWOO} \cite{rewoo}, \textbf{\textsc{FireAct}} \cite{fireact}.
We detail each baseline in Appx.~\ref{app:baseline}.
To ensure fairness, we maintain an equal training trajectory volume of 200 for \textsc{FireAct} and {\ours} (200 synthesized data).
As Reflexion provides answer correctness labels during reflection but other methods including {\ours} do not, we test all the other methods twice and choose the correct one for evaluation.
For all the prompt-based baselines, we uniformly provide two examples in the prompt.

\paragraph{Training Setups.}
We fine-tune all our models with LoRA \cite{lora} in the format proposed in Alpaca \cite{alpaca}.
All the training and inference experiments are conducted on 8 V100 GPUs within 16 hours.
We detail the hyper-parameters for training in Appx.~\ref{app:hp}.

\section{Results}
\paragraph{Compare to Prompt-based Agent Learning Baselines.}
As shown in Tab.~\ref{tab:main_results}, the Mistral-7B and Llama-\{13,70\}B models consistently outperform various prompt-based baselines.
The Llama-70B model even surpasses the agent performance of GPT-3.5-Turbo, achieving a rise of \daulg{3.77\%} on HotpotQA and \daulg{6.39\%} on ScienceQA.
Therefore, whether in a single-agent or multi-agent architecture, prompt-based methods relying on few-shot demonstrations fail to precisely customize the behavior of the agent, which is also supported by the fact that \textsc{FireAct} widely outperforms \textsc{ReAct} and BOLAA in the context of iterative planning.

\paragraph{Compare to Fine-tuning-based Agent Learning Baselines.}
Further focusing on \textsc{FireAct} in Tab.~\ref{tab:main_results}, despite the aid of GPT-4, \textsc{FireAct}'s approach of assigning the entire planning task to a single model proves to be burdensome.
As a result, its performance on ScienceQA even falls short compared to the prompt-based global planning method, Chameleon.
{\ours} decouples the planning process and reaches a clear \textit{division-of-labor} among sub-agents for group planning, resulting in an improvement than \textsc{FireAct}, with \daulg{5.77\%} on HotpotQA and \daulg{6.67\%} on ScienceQA with Llama-70B model.
Additionally, {\ours} achieves self-planning without relying on closed-source models and large-scale labeled datasets, which paves the way for automatic agent learning with open-source models from scratch.
In ablation study (\S\ref{para:ablation}) and human evaluation (\S\ref{para:human}), we will further validate that the quality of trajectories synthesized by {\ours} is not inferior to \textsc{FireAct} trained on trajectories synthesized using GPT-4.

\begin{figure*}[t!]
    \centering
    \resizebox{0.95\textwidth}{!}{
    \includegraphics{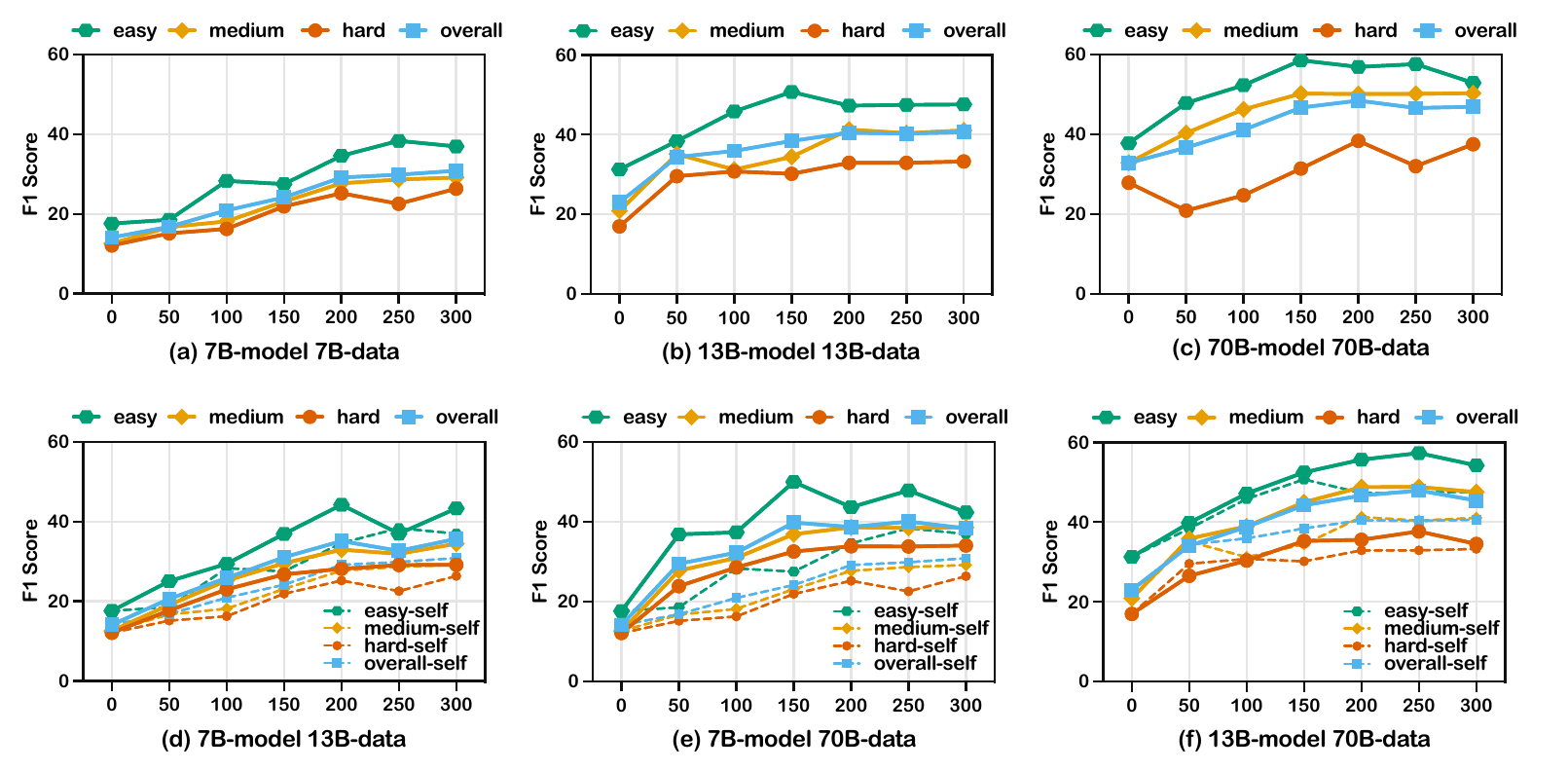}}
    \caption{
    \textbf{Performance of {\ours} on HotpotQA with different training data scales.}
    The \{7,13,70\}B represents Llama-2-\{7,13,70\}B-chat models respectively.
    (a-c) shows the results of the model trained on self-synthesized trajectories.
    (d-f) represents the results of the model trained on trajectories synthesized by a stronger model, where the dashed line is the baseline trained on self-synthesized trajectories.
    }
    \label{fig:ft_num}
\end{figure*}

\begin{figure*}[t!]
    \centering
    \resizebox{0.95\textwidth}{!}{
    \includegraphics{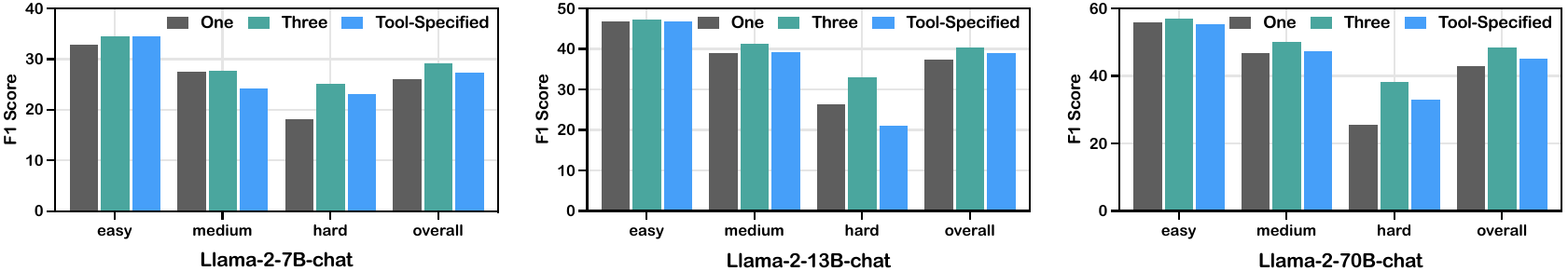}}
    \caption{
    \textbf{Performance of {\ours} on HotpotQA based on different degrees of labor division.}
    \textbf{\textit{One}} is training a single model with all the differentiated data.
    \textbf{\textit{Three}} represents the differentiation into three agents: plan, tool, and reflect.
    \textbf{\textit{Tool Specified}} indicates further differentiating the tool-agent with one tool, one agent.
    }
    \label{fig:degree}
\end{figure*}

\begin{table}[t!]
\centering
\renewcommand\arraystretch{1}
\scalebox{.86}{
\begin{tabular}{rcc}
\hline
\toprule
 & \multicolumn{1}{c}{\textbf{HotpotQA}} & \multicolumn{1}{c}{\textbf{ScienceQA}} \\
\Xhline{1px}
\textbf{\ours} & \multicolumn{1}{c}{48.47} & \multicolumn{1}{c}{78.61}  \\
\textit{- reflection} & 45.66$_{\downarrow2.81}$ & 75.28$_{\downarrow3.33}$ \\
\textit{- multi} & 42.81$_{\downarrow5.66}$ & 69.72$_{\downarrow8.89}$ \\
\textit{- fine-tuning} & 32.84$_{\downarrow15.63}$ & 61.94$_{\downarrow16.67}$ \\
\textit{- filtering} & 32.51$_{\downarrow15.96}$ & 59.17$_{\downarrow19.44}$ \\
\bottomrule
\hline
\end{tabular}
}
\caption{
\textbf{Approach ablations of {\ours}.}
\textbf{\textit{- reflection}} symbolizes removing the reflect-agent in {\ours}.
\textbf{\textit{- multi}} denotes feeding all the differentiated data into one model for fine-tuning.
\textbf{\textit{- fine-tuning}} indicates zero-shot prompt planning with the three agents defined in {\ours}.
\textbf{\textit{- filtering}} represents self-differentiation on all the trajectories generated in zero-shot planning without filtering wrong cases.
}
\label{tab:ablation}
\end{table}

\paragraph{Single-agent Learning vs. Multi-agent Learning.}
Under identical settings, multi-agent architectures generally exhibit better performance than single-agent (\textsc{ReAct} vs. BOLAA, \textsc{FireAct} vs. {\ours}), which aligns with Simon's theory of bounded rationality.
Seemingly contrary to expectations, despite being a single-agent architecture, Chameleon outperforms BOLAA (even \textsc{FireAct} on ScienceQA).
However, we analyze that this can be attributed to the way it leverages tools.
In Chameleon, the process of deciding tool parameters is considered a form of tool invocation, and specialized few-shot prompts are designed to guide the model through this process.
From this aspect, Chameleon, despite nominally a single-agent architecture, exhibits features resembling a multi-agent one, which does not contradict our initial conclusion.
Indeed, we can also explain from the perspective of optimizing objectives.
Another well-known principle, Goodhart's Law \citep{goodhart}, states that \textit{``When a measure becomes a target, it ceases to be a good measure''.}
This implies that optimizing one objective on the same agent will inevitably harm other optimization objectives to some extent. 
Therefore, it is not optimal to optimize all objectives on a single agent, and a multi-agent architecture happens to address this issue.
However, we analyze in \S\ref{para:degree} that excessive fine-grained \textit{division-of-labor} is not the best approach.

\begin{figure*}[t!]
    \centering
    \resizebox{0.9\textwidth}{!}{
    \includegraphics{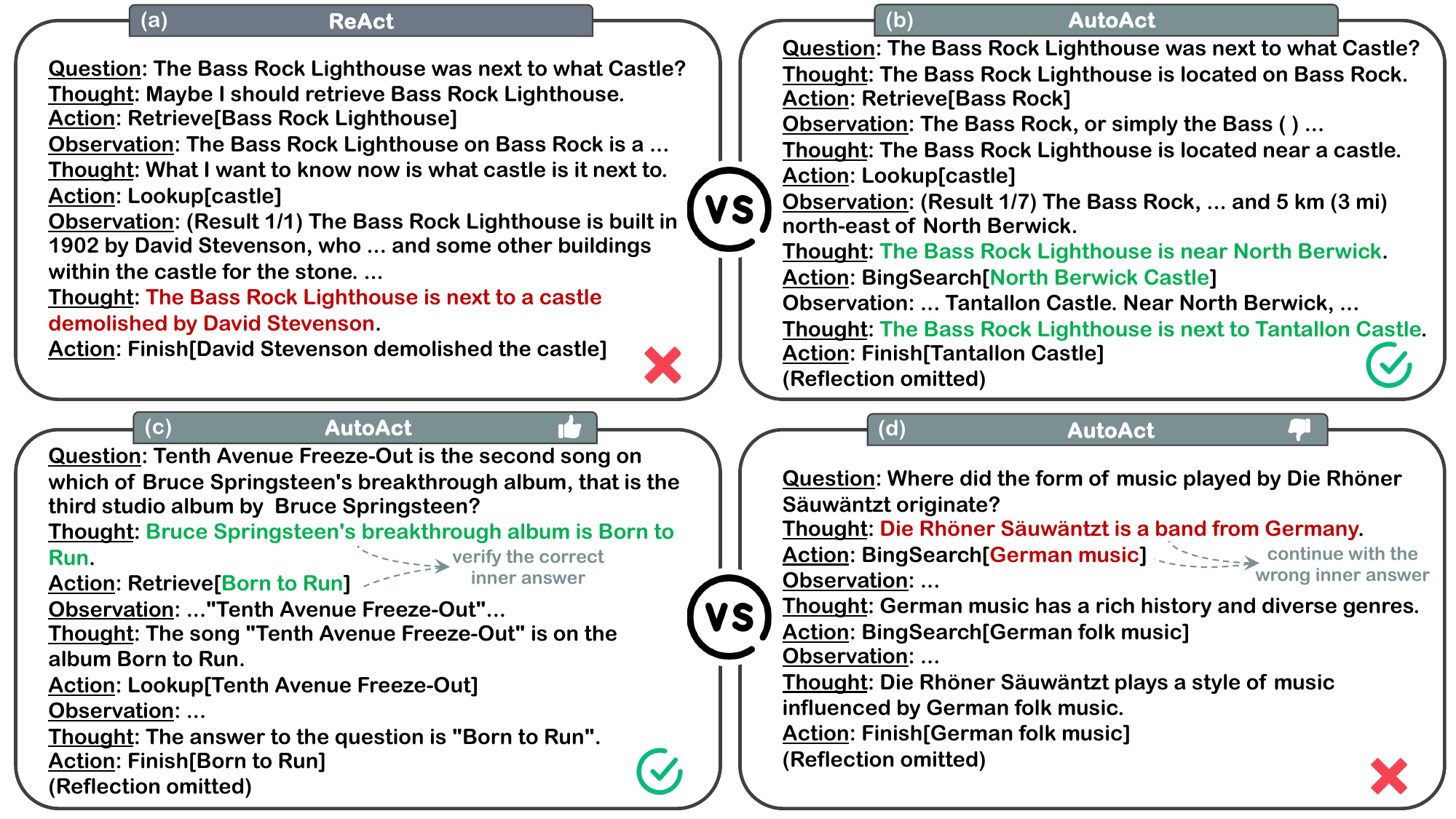}}
    \caption{
    \textbf{Case study} on HotpotQA.
    {\ours} (b) successfully addresses the failure in \textsc{ReAct} (a) by employing a more scientific combination of tools and making more accurate tool invocations.
    With more planning rounds, {\ours} (c) can validate its inner answers by continuing more rounds of self-verification.
    While this can also lead to a longer context, gradually deviating {\ours} (d) from the original question.
    }
    \label{fig:case}
\end{figure*}

\begin{figure}[t!]
    \centering
    \resizebox{0.42\textwidth}{!}{
    \includegraphics{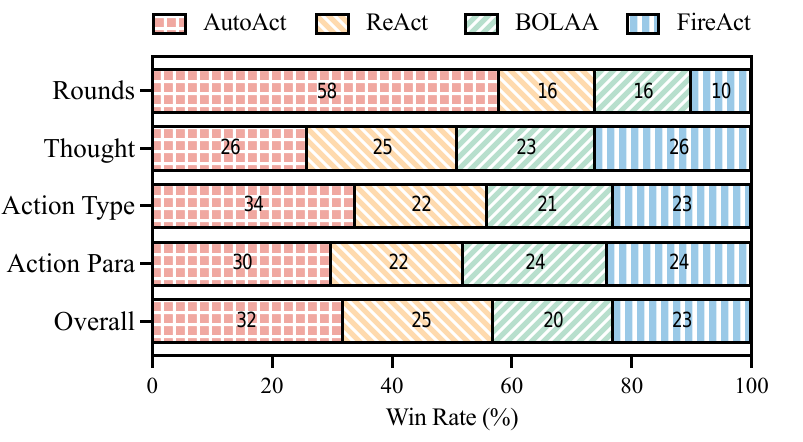}}
    \caption{
    \textbf{Human evaluation of trajectories} generated by Llama-2-70B-chat on HotpotQA.
    We compare the number of planning rounds, the logical correctness of thoughts, action types, action parameters, and the overall coherence of each trajectory.
    The figure above displays the \textbf{Win Rate} of each method in each aspect.
    }
    \label{fig:human}
\end{figure}

\paragraph{Approach Ablations.}
\label{para:ablation}
Tab.~\ref{tab:ablation} presents the performance of {\ours} on the Llama-70B model after removing certain key processes.
It can be observed that the least impactful removal is the \textit{- reflect}.
We investigate that in the zero-shot scenario, the model tends to be over-confident in its answers \revise{(as also confirmed in \citet{self-correct})}.
It typically only recognizes its errors when there are obvious formatting mistakes or significant repetitions in the planning process.
Consistent with previous findings, the removal of the \textit{- multi} agents leads to a noticeable decrease in performance.
A more exciting discovery is that the results of \textit{- multi} are comparable to those of \textsc{FireAct}.
This indirectly suggests that the trajectory quality generated by the 70B model may be no worse than that of GPT-4.
As expected, the performance deteriorates after \textit{- fine-tuning}, which once again confirms the previous conclusion.
To demonstrate the necessity of filtering out planning error data, we specifically remove the filtering process (\textit{- filtering}) to examine the performance of {\ours}.
The results indicate that the damage caused by training on unfiltered data is even greater than that of \textit{- fine-tuning}.

\section{Analysis}
\label{sec:analysis}
\paragraph{Larger training data scale does not necessarily mean better results.}
\label{para:ft_num}
We evaluate the influence of different training data scales on the performance of self-planning with Llama-\{7,13,70\}B models on HotpotQA in Fig.~\ref{fig:ft_num} (a-c).
It can be observed that the overall performance of different models goes to stability with minimal waves once the data scale exceeds 200.
We speculate that this may be due to the limited ability of naive self-instruct to boost internal knowledge of the language model.
As the training data increases, the knowledge which can be extracted through self-instruct decreases.
Despite our efforts to filter out duplicate data, the mindless increase can inevitably lead to a significant surge in similar data, which undermines the benefits of increasing the data scale and makes it challenging to improve model performance or even leads to over-fitting.
To further confirm the role of training data, we decouple the models from the training data and evaluate their training results on trajectories synthesized by stronger models.
From Fig.~\ref{fig:ft_num} (d-f), we can see consistent conclusions with previous findings.
Therefore, maximizing the diversity of the synthesized data in the database may be a key improvement direction for {\ours} and we leave this for our future work.
\revise{We can also observe from Fig.~\ref{fig:ft_num} (d-e) that the larger the model, the higher the quality of the synthesized data, as the performance of the 7B model shows a gradual increase on self, 13B, and 70B synthesized data.}

\paragraph{Moderate division-of-labor benefits group planning performance.}
\label{para:degree}
To explore the impact of different granularity of self-differentiation, we further subdivide the tool agent, assigning dedicated agents to manipulate each specific tool.
We compare the performance of \textit{One} agent, \textit{Three} agents ({\ours}), and the \textit{Tool-Specified} setting on HotpotQA in Fig.~\ref{fig:degree}.
It can be observed that excessive differentiation (\textit{Tool-Specified}) not only fails to achieve better results but can sometimes even be less effective than not differentiating (\textit{One}) at all.
This is consistent with the findings in \citet{trice} which indicate that multi-tool joint learning often outperforms single-tool individual learning.
Moreover, it appears that the performance loss of tool-specific agents compared to {\ours} is more significant on harder problems.
This is because challenging problems typically require more planning steps and higher levels of collaboration among tools.
By unifying tool invocations under one agent, it becomes possible to effectively learn the interconnectedness between tools, thereby compensating for potential information gaps arising from using tool-specific agents.
Note the difference from \citet{more/agents/allyouneed}, here we are discussing the granularity of division-of-labor among agents with different responsibilities, rather than the voting quantity among mutually equal agents.

\paragraph{Human Evaluation.}
\label{para:human}
To get a deeper understanding of the quality of trajectories generated by different methods, we manually compare them from the number of planning rounds, the logical correctness of thoughts, action types, action parameters, and overall coherence.
The detailed human evaluation process can be found in Appx.~\ref{app:human}.
The evaluation results are depicted in Fig.~\ref{fig:case}\&\ref{fig:human}.
We can observe a clear advantage for {\ours} over other methods in the action type and action parameters.
This indicates that decoupling the missions of planning and tool invocation can lead to better performance for both, alleviating the overwhelming pressure on a single agent.
A more intuitive comparison can be observed in Fig.~\ref{fig:case} (a-b).
{\ours} successfully addresses the failure in \textsc{ReAct} by employing a more scientific combination of tools and making more accurate tool invocations.
Furthermore, {\ours} tends to consume more planning rounds than other methods (the specific average planning rounds is in Appx.~\ref{app:planning_rounds}).
This allows {\ours} to perform better on harder problems.
However, this characteristic can be a double-edged sword when it comes to simple problems.
A surprising aspect is that {\ours} can validate its inner answers by continuing more rounds of verification (Fig.~\ref{fig:case} (c)). 
But this can also lead to a longer context, gradually deviating {\ours} from the original question (Fig.~\ref{fig:case} (d)).

\section{Related Work}

\paragraph{LLM-Powered Agents.}
The rise of LLMs has positioned them as the most promising key to unlocking the door to Artificial General Intelligence (AGI), providing robust support for the development of LLM-centered AI agents \cite{renda/agent/survey,fudan/agent/survey,wang2023describe,wang2023jarvis}.
Related works focus primarily on agent planning \cite{react,llm-planner,fireact}, external tools harnessing \cite{gorilla,trice,toolllm}, collective intelligence among multi-agents \cite{debate,bolaa,round/table}, etc.
However, despite their success, existing methods still face two major troubles.
\textbf{Firstly}, most agents heavily rely on prompts for customization, which makes it difficult to precisely tailor the behavior of the agent, resulting in unexpected performance at times.
\textbf{Secondly}, each agent is compelled to master all skills, making it challenging for the agent to achieve expertise in every domain.
In response, our approach leverages a proper \textit{division-of-labor} strategy and fine-tuning each sub-agent to equip different agents with distinct duties.
These agents collaborate to accomplish tasks orderly and effectively.

\paragraph{Agent Fine-Tuning.}
Despite the vast interest in LLM-powered agents, the construction of agents through fine-tuning has received limited attention.
Most early works concentrate on fine-tuning to optimize the model's reasoning capabilities \cite{generated/commonsense,specialize/small} or tool proficiency \cite{gorilla,trice,toolllm}.
Recently, more works have emphasized endowing open-source LLMs with agent capabilities through fine-tuning \cite{fireact,agenttuning,lumos,small/llms/weak/tool}.
However, these works suffer from at least one of the following issues:
\textbf{\textit{i)}} the requirement of one single model to be a generalist,
\textbf{\textit{ii)}} the need for a large amount of annotated data,
\textbf{\textit{iii)}} the need for trajectory annotation of closed-source models.
Our approach enables the \textsc{Meta-Agent} to synthesize trajectories and achieve a \textit{division-of-labor} strategy in a zero-shot manner, without relying on closed-source models.

\section{Conclusion and Future Work}
In this paper, we propose {\ours}, an automatic agent learning framework for QA that does not rely on large-scale annotated data and synthetic trajectories from closed-source models, while alleviating the pressure on individual agents by explicitly dividing the workload.
Interesting future directions include:
\textbf{\textit{i)}} expanding {\ours} to more realistic task scenarios \citep{virtualhome,webarena,travelplanner},
\textbf{\textit{ii)}} boosting more knowledge via self-instruct (as analyzed in \S\ref{para:ft_num}),
\textbf{\textit{iii)}} iteratively enhancing synthetic trajectories via self-improvement \citep{lmsi,rest/meets/react}.

\section*{Limitations}
In this paper, we focus on constructing an automatic agent learning framework dubbed {\ours}.
Despite our best efforts, this paper may still have some remaining limitations.

\paragraph{Tasks.}
In this paper, we mainly focus on complex question-answering tasks.
However, there are many other more complex interactive scenarios, including web \cite{webshop,webarena}, household \cite{virtualhome,alfworld}, traveling \cite{travelplanner}, robotics \cite{saycan}, etc.
For example, we have investigated the use of \textsc{Meta-Agent} performing random explorations \citep{world_model,bagel} in virtual environments to replace the process of task and trajectory synthesis through self-instruct and zero-shot planning.
We plan to conduct further research on applying {\ours} to a wider range of tasks based on this in the future.

\paragraph{Boosting Knowledge via Self-Instruct.}
As analyzed in \S\ref{sec:analysis}, the planning performance of {\ours} can be limited by the model's ability to access internal knowledge through self-instruct.
While the current phenomenon allows us to achieve lightweight self-differentiation in terms of parameters and data, it is still necessary to research how to enrich knowledge as much as possible within the constraints of limited data.

\paragraph{Self-Improvement.}
Recent research has shed light on self-improvement techniques that enhance LLMs by iteratively training them on self-synthesized data \cite{star,lmsi,rest,rest/meets/react}.
This approach allows the model to continually learn and refine its performance on its own.
Our approach also involves training on self-synthesized data and we believe that further using the iterative thinking of self-improvement will significantly enhance the performance of our method.

\section*{Ethics Statement}
This research was conducted with the highest ethical standards and best practices in research.
All our experiments use publicly available datasets (as detailed in \S\ref{para:task}), avoiding ethical concerns related to privacy, confidentiality, or misuse of personal biological information.
The human evaluation process (as detailed in Appx.~\ref{app:human}) was carried out strictly with fairness and transparency.
Consequently, this research is free from any ethical concerns.

\section*{Acknowledgements}
We would like to express our sincere gratitude to the anonymous reviewers for their thoughtful and constructive feedback.
This work was supported by the National Natural Science Foundation of China (No. 62206246), the Fundamental Research Funds for the Central Universities (226-2023-00138), Zhejiang Provincial Natural Science Foundation of China (No. LGG22F030011), Yongjiang Talent Introduction Programme (2021A-156-G), Tencent AI Lab Rhino-Bird Focused Research Program (RBFR2024003), and Information Technology Center and State Key Lab of CAD\&CG, Zhejiang University. 

\bibliography{custom}
\bibliographystyle{acl/acl_natbib}

\appendix

\section{Comparison with Related Works}
\label{app:compare}
See Tab.~\ref{tab:compare}

\begin{table*}[t!]
\centering
\renewcommand\arraystretch{1}
\scalebox{0.63}{
\begin{tabular}{lccccccc}
\hline
\toprule
\textbf{Method} & \textbf{\makecell{Data\\Acquisition}} & \textbf{\makecell{Trajectory\\Acquisition}} & \textbf{Planning} & \textbf{Multi-Agent} & \textbf{Fine-Tuning} & \textbf{Generality} & \textbf{Reflection} \\
\Xhline{1px}
\textsc{ReAct} \cite{react} & User & Prompt & Iterative & \colorxmark & \colorxmark & \colorcmark & \colorxmark \\
Reflexion \cite{reflexion} & User & Prompt & Iterative & \colorxmark & \colorxmark & \colorcmark & \colorcmark \\
Camel \cite{camel} & User & Prompt & Iterative & \colorcmark & \colorxmark & \colorcmark & \colorxmark \\
Chameleon \cite{chameleon} & User & Prompt & Global & \colorxmark & \colorxmark & \colorcmark & \colorxmark \\
HuggingGPT \cite{hugginggpt} & User & Prompt & Global & \colorxmark & \colorxmark & \colorcmark & \colorxmark \\
AutoGPT \cite{autogpt} & User & Prompt & Iterative & \colorxmark & \colorxmark & \colorcmark & \colorcmark \\
BOLAA \cite{bolaa} & User & Prompt & Iterative & \colorcmark & \colorxmark & \colorcmark & \colorxmark \\
AgentVerse \cite{agentverse} & User & Prompt & Iterative & \colorcmark & \colorxmark & \colorcmark & \colorxmark \\
Agents \cite{agents} & User & Prompt & Iterative & \colorcmark & \colorxmark & \colorcmark & \colorxmark \\
AgentTuning \cite{agenttuning} & Benchmark & GPT-4 & Iterative & \colorxmark & \colorcmark & \colorxmark & \colorxmark \\
\textsc{FireAct} \cite{fireact} & Benchmark & GPT-4 & Iterative & \colorxmark & \colorcmark & \colorxmark & \colorcmark \\
Lumos \cite{lumos} & Benchmark & Benchmark + GPT-4 & Both & \colorcmark & \colorcmark & \colorxmark & \colorxmark \\
\textbf{\ours}\ (ours) & User + Self-Instruct & Self-Planning & Iterative & \colorcmark & \colorcmark & \colorcmark & \colorcmark \\
\bottomrule
\hline
\end{tabular}
}
\caption{\textbf{Comparison of related works.}
\textbf{Data} and \textbf{Trajectory Acquisition}s refer to the way for obtaining training data and trajectories.
\textbf{Planning} represents the way of planning, parted based on whether each step's action is determined globally or iteratively.
\textbf{Multi-Agent} indicates whether the framework contains multi-agent.
\textbf{Fine-Tuning} stands for whether the method is a fine-tuning-based agent learning framework.
\textbf{Generality} signifies whether the method is applicable to various tasks.
\textbf{Reflection} denotes whether the planning process incorporates reflection.}
\label{tab:compare}
\end{table*}

\section{Baselines and Training Setups}
\paragraph{Baselines.}
\label{app:baseline}
We choose the open-source Llama-2 models \cite{llama2} and Mistral-7B \cite{mistral} as the backbones of our \textsc{Meta-Agent} and sub-agents.
The compared baselines are as follows:
1) \textbf{CoT} \cite{cot}, the naive Chain-of-Thought reasoning method.
2) \textbf{\textsc{ReAct}} \cite{react}, a well-known single-agent framework based on few-shot learning that performs planning and action iteratively.
3) \textbf{Chameleon} \cite{chameleon}, another few-shot single-agent framework that performs planning before action.
4) \textbf{Reflexion} \cite{reflexion}, a single-agent framework to reinforce language agents through linguistic feedback.
5) \textbf{BOLAA} \cite{bolaa}, a multi-agent framework that customizes different agents through prompts.
6) \textbf{ReWOO} \cite{rewoo}, a multi-agent framework that decouples reasoning from observations.
7) \textbf{\textsc{FireAct}} \cite{fireact}, a single-agent framework with fine-tuning on diverse kinds of trajectories generated by GPT-4 \cite{gpt-4}.
8) \textbf{GPT-3.5-Turbo} \cite{gpt-3.5}.
To ensure fairness, we maintain an equal training trajectory volume of 200 for \textsc{FireAct} and {\ours} (200 synthesized data).
As Reflexion provides answer correctness labels during reflection but other methods including {\ours} do not, we test all the other methods twice and choose the correct one for evaluation.
For all the prompt-based baselines, we uniformly provide two examples in the prompt.

\paragraph{Training Setups.}
\label{app:hp}
We fine-tune all our models with LoRA \cite{lora} in the format proposed in Alpaca \cite{alpaca}.
Our fine-tuning framework leverages FastChat \cite{fastchat} using DeepSpeed \cite{deepspeed}.
We detail the hyper-parameters for training in Tab.~\ref{tab:hp}.

\begin{table*}[t!]
    \centering
    \renewcommand\arraystretch{1}
    \scalebox{1.}{
    \begin{tabular}{ccc}
    \toprule
    \textbf{Name} &  \textbf{Mistral-7B\&Llama-2-\{7,13\}B-chat} & \textbf{Llama-2-70B-chat} \\
    \Xhline{1px}
    lora\_r & 8 & 8 \\
    lora\_alpha & 16 & 16 \\
    lora\_dropout & 0.05 & 0.05 \\
    lora\_target\_modules & q\_proj, v\_proj & q\_proj, v\_proj \\
    model\_max\_length & 4096 & 4096 \\
    per\_device\_batch\_size & 2 & 2 \\
    gradient\_accumulation\_steps & 1 & 1 \\
    warmup\_ratio & 0.03 & 0.03 \\
    epochs & 5 & 3 \\
    batch size & 4 & 1 \\
    learning rate & 1e-4 & 1e-4 \\
    \bottomrule
    \end{tabular}
    }
    \caption{Detailed hyper-parameters we use for training.}
    \label{tab:hp}
\end{table*}

\section{Detailed Process of Human Evaluation}
\label{app:human}
To get a deeper understanding of the capability of {\ours}, we manually compare the quality of trajectories generated by different methods from five aspects.
We ask five NLP volunteers to individually select the optimal trajectories generated by all methods in terms of the number of planning rounds, the logical correctness of thoughts, action types, action parameters, and overall coherence.
The final results are determined based on major votes.
During the evaluation, it is hidden for the evaluators of the correspondence between the trajectories and the methods.
We delete the reflection-related parts from the trajectories generated by {\ours} and randomly shuffle the order of trajectories of each method in each data to minimize the potential bias as much as possible.

\section{Average Planning Rounds}
\label{app:planning_rounds}
We compare the planning rounds of {\ours} with various baselines.
The win rate of each method is listed in Fig.~\ref{fig:human} and comprehensive analysis can be found in \S\ref{para:human}.
Here we present the average planning rounds of various methods on HotpotQA with Llama-2-70B-chat in Tab.~\ref{tab:planning_rounds}.
Note that to maintain fairness, we exclude the planning steps related to reflection of {\ours}.

\begin{table}[t!]
    \centering
    \renewcommand\arraystretch{1}
    \scalebox{1.}{
    \begin{tabular}{lccc}
    \toprule
    \textbf{Method} & \textbf{Easy} & \textbf{Medium} & \textbf{Hard} \\
    \Xhline{1px}
    \textsc{ReAct} & 3.83 & 4.02 & 4.13 \\
    BOLAA & 3.60 & 3.76 & 3.96 \\
    \textsc{FireAct} & 3.01 & 3.17 & 3.70 \\
    \textbf{\ours} & 4.62 & 4.73 & 4.96 \\
    \bottomrule
    \end{tabular}
    }
    \caption{Average planning rounds of various methods on HotpotQA with Llama-2-70B-chat.}
    \label{tab:planning_rounds}
\end{table}

\section{Task Information}
\label{app:task}
\textbf{Task Name}: \textbf{HotpotQA}\\
\textbf{Task Description}: This is a question-answering task that includes high-quality multi-hop questions.
It tests language modeling abilities for multi-step reasoning and covers a wide range of topics. Some questions are challenging, while others are easier, requiring multiple steps of reasoning to arrive at the final answer.\\
\textbf{Task Data Examples}:\\
 \underline{Question}: From 1969 to 1979, Arno Schmidt was the executive chef of a hotel located in which neighborhood in New York? \\
 \underline{Answer}: Manhattan \\
 \\
 \underline{Question}: Are both Shangri-La City and Ma'anshan cities in China? \\
 \underline{Answer}: yes \\
 \\
\textbf{Task Name}: \textbf{ScienceQA}\\
\textbf{Task Description}: This is a multimodal question-answering task that necessitates a model to utilize tools for transforming image information into textual data. Simultaneously, this task incorporates substantial background knowledge, requiring the language model to acquire external information to enhance its comprehension of the task.\\
\textbf{Task Data Examples}:\\
\underline{Question}: Which of these states is the farthest north?\\
\underline{Options}: (A) West Virginia (B) Louisiana (C) Arizona (D) Oklahoma\\
\underline{Caption}: An aerial view of a painting of a forest. \\
\underline{Answer}: A. West Virginia \\
\\
\underline{Question}: Identify the question that Tom and Justin's experiment can best answer. \\
\underline{Context}: The passage below describes an experiment. Read the passage and then follow the instructions below. Tom placed a ping pong ball in a catapult, pulled the catapult's arm back to a 45 angle, and launched the ball. Then, Tom launched another ping pong ball, this time pulling the catapult's arm back to a 30 angle. With each launch, his friend Justin measured the distance between the catapult and the place where the ball hit the ground. Tom and Justin repeated the launches with ping pong balls in four more identical catapults. They compared the distances the balls traveled when launched from a 45 angle to the distances the balls traveled when launched from a 30 angle. Figure: a catapult for launching ping pong balls.\\
\underline{Options}: (A) Do ping pong balls stop rolling along the ground sooner after being launched from a 30-angle or a 45-angle? (B) Do ping pong balls travel farther when launched from a 30-angle compared to a 45-angle? \\
\underline{Caption}: A wooden board with a wooden head on top of it. \\
\underline{Answer}: B. Do ping pong balls travel farther when launched from a 30 angle compared to a 45 angle? \\
\\

\section{Tool Library}
\label{app:tool}
To facilitate our agents in automatic task planning, we provide a comprehensive tool library that contains 15 commonly used tools for various complex question-answering tasks.
A part of our tools and their corresponding information can be found in Tab.~\ref{tab:tool_library}.
Users can also have the option to expand the tool library according to their specific needs, allowing for more flexible utilization.

\begin{table*}[t!]
    \centering
    \renewcommand\arraystretch{1}
    \scalebox{1.}{

    \begin{tabular}{cp{5cm}p{5cm}}
    \toprule
    \textbf{Name} & \textbf{Definition} & \textbf{Usage} \\
    \Xhline{1pt}
    BingSearch &  BingSearch engine can search for rich knowledge on the internet based on keywords, which can compensate for knowledge fallacy and knowledge outdated. &  BingSearch[query], which searches the exact detailed query on the Internet and returns the relevant information to the query. Be specific and precise with your query to increase the chances of getting relevant results. For example, Bingsearch[popular dog breeds in the United States] \\
    \hline
    Retrieve & Retrieve additional background knowledge crucial for tackling complex problems. It is especially beneficial for specialized domains like science and mathematics, providing context for the task & Retrieve[entity], which retrieves the exact entity on Wikipedia and returns the first paragraph if it exists. If not, it will return some similar entities to retrieve. For example, Retrieve[Milhouse] \\
    \hline
    Lookup & A Lookup Tool returns the next sentence containing the target string in the page from the search tool, simulating Ctrl+F functionality on the browser. & Lookup[keyword], which returns the next sentence containing the keyword in the last passage successfully found by Retrieve or BingSearch. For example, Lookup[river]. \\
    \hline
    Image2Text & Image2Text is used to detect words in images convert them into text by OCR and generate captions for images. It is particularly valuable when understanding an image semantically, like identifying objects and interactions in a scene. & Image2Text[image], which generates captions for the image and detects words in the image. You are recommended to use it first to get more information about the image to the question. If the question contains an image, it will return the caption and OCR text, else, it will return None. For example, Image2Text[image].\\
    \hline
    Text2Image & Text2Image Specializes in converting textual information into visual representations, facilitating the incorporation of textual data into image-based formats within the task. & Text2Image[text], which generates an image for the text provided by using multimodal models. For example, Text2Image[blue sky]\\
    \hline
        ...... & ......& ......\\
    \hline
    Code Interpreter & Code Interpreter is a tool or software that interprets and executes code written in Python. It analyzes the source code line by line and translates it into machine-readable instructions or directly executes the code and returns Execution results& Code[python], which interprets and executes Python code, providing a line-by-line analysis of the source code and translating it into machine-readable instructions. For instance, Code[print("hello world!")]\\
    \bottomrule
    \end{tabular}
    }
    \caption{Part of our tool library.}
    \label{tab:tool_library}
\end{table*}

\section{Prompt}

\subsection{Prompt for Self-Instruct}
\label{app:prompt_self_instruct}

See Tab.~\ref{tab:prompt_self_instruct}.

\begin{table*}[t!]
    \centering
    \renewcommand\arraystretch{1}
    \scalebox{1.}{
    \begin{tabular}{p{14cm}}
    \large{\textbf{Prompt for Self-Instruct}}\\
    \hline
    I want you to be a QA pair generator to generate high-quality questions for use in Task described as follows:\par
    Task Name: \textbf{[task\_name]}\par
    Task Description: \textbf{[task\_description]}\par
    Here are some Q\&A pair examples from the Task:\par
    \textbf{[QA\_pairs]}\par 
    Modeled on all the information and examples above, I want you to generate new different \textbf{[gen\_num\_per\_round]} Question-Answer pairs that cover a wide range of topics, some of which are difficult, some of which are easy, and require multiple steps of reasoning to get to the final answer. The format is like below:\par
    \textbf{[one\_example]}\par
    \end{tabular}
    }
    \caption{Prompt used for self-instruct.}
    \label{tab:prompt_self_instruct}
\end{table*}

\subsection{Prompt for Tool Selection}
\label{app:prompt_tool}

See Tab.~\ref{tab:prompt_tool}.

\begin{table*}[t!]
    \centering
    \renewcommand\arraystretch{1}
    \scalebox{1.}{
    \begin{tabular}{p{14cm}}
    \large{\textbf{Prompt for Automatic Tool Selection}}\\
    \hline
    To successfully complete a complex task, the collaborative effort of three types of agents is typically required: \par
    1. Plan Agent. This agent is used to plan the specific execution process of the benchmark, solving a given task by determining the order in which other expert language models are invoked;\par
    2. Tool Agent. This agent is employed to decide how to use a specific tool when addressing a task. Tools encompass interactive tools within the task environment as well as external tools or models. The Tool Agent includes various tools that can be flexibly chosen;\par
    3. Reflect Agent. This agent reflects on historical information and answers to assess whether the response aligns with the provided query.\par
    Above all, the Tool Agent includes many tools that can be flexibly selected. Now your task is to select 3 tools from the Tool Library for solving a given task. Note that all tools are based on language models, and their inputs and outputs must be text. You only need to provide the names and descriptions of the tools in order, without any additional output.\\
    \textbf{Task Prompt Template} \\
    The following is the given task name and description, and you need to choose 3 corresponding tools from the Tool Library according to the above rules in the format of one line, one tool.\par
    Task Name: \textbf{[task\_name]}\par
    Task Description: \textbf{[task\_description]}\par
    Tool Library: \textbf{[list\_of\_tools]}\par
    \end{tabular}
    }
    \caption{Prompt used for automatic tool selection.}
    \label{tab:prompt_tool}
\end{table*}

\subsection{Prompt for Trajectories Synthesis}
\label{app:prompt_traj_syn}

See Tab.~\ref{tab:prompt_traj_syn}.

\begin{table*}[t!]
    \centering
    \renewcommand\arraystretch{1}
    \scalebox{1.}{
    \begin{tabular}{p{14cm}}
    \large{\textbf{Prompt for Trajectories Synthesis}}\\ 
    \hline
    I expect you to excel as a proficient question answerer in the task.\par
    Task Name: \textbf{[task\_name]}\par
    Task Description: \textbf{[task\_description]}\par
    Solve a question-answering task with interleaving Thought, Action, and Observation steps. Thought can reason about the current situation, and Action can be \textbf{[action\_num]} types:\par
    \textbf{list of action selected from automatic tool selection [name, definition , usage]}\par
    Question: \textbf{[question]}\textbf{[scratchpad]}\par
    \end{tabular}
    }
    \caption{Prompt used for trajectories synthesis.}
    \label{tab:prompt_traj_syn}
\end{table*}

\section{Database Cases}
\label{app:database}

\textbf{HotpotQA}: \\
Question: The deepest part of the ocean, is located in which ocean? \\
Answer: The Pacific Ocean\\
\\
Question: The famous scientist who discovered gravity, lived in which century?\\
Answer: 17th century\\
\\
Question: The first successful flight of a power was made by which inventor?\\
Answer: The Wright brothers\\
\\
Question: The highest mountain peak in the solar system is located on which planet?\\
Answer: Mars\\
\\
Question: In the novel "Pride and Prejudice", what is the name of Mr. Darcy's estate in Derbyshire, England?\\
Answer: Pemberley\\
\\
\textbf{ScienceQA}: \\
Question: Which of the following is a type of renewable energy?\\
Options: (A) Coal (B) Oil (C) Natural gas (D) Solar power\\
Caption: A picture of a solar cell\\
Answer: D. Solar power\\
\\
Question: Which of the following is the term for the process by which the Earth's weather patterns are influenced by the movement of air in the atmosphere?\\
Options: (A) Weathering (B) Erosion (C) Deposition (D) Atmospheric circulation\\
Caption: An image of air currents in the atmosphere\\
Answer: D. Atmospheric circulation\\
\\
Question: Which of the following is a type of chemical reaction that involves the transfer of electrons between atoms?\\
Options: (A) Combustion (B) Photosynthesis (C) Respiration (D) Electrolysis\\
Caption: An image of a battery\\
Answer: D. Electrolysis\\
\\
Question: Which of the following is an example of a type of weather phenomenon that occurs when warm air rises and cool air sinks?\\
Options: (A) Thunderstorms (B) Hurricanes (C) Fog (D) Fronts\\
Caption: An image of a front\\
Answer": D. Fronts\\
\\
Question: Which of the following is the term for the process by which water is purified through the use of microorganisms that consume organic matter?\\
Options: (A) Filtration (B) Sedimentation (C) Biodegradation (D) Disinfection\\
Caption: An image of a water treatment plant\\
Answer: C. Biodegradation

\section{Training Data Example}
\label{app:training_data_example}
Here we give an example of the training data for each sub-agent.\\
\\
\textbf{Plan-Agent (generate Thought)}:\\
\textbf{Input}:\\
(format requirements)
(tool usage instructions)\\
\underline{Question}: The first human-made object to land on the moon, in 1969, was which spacecraft?\\
\underline{Thought}: I should first search the Moon landing history.\\
\underline{Action}: BingSearch[moon landing spacecraft]\\
\underline{Observation}: A Moon landing or lunar landing is the arrival of a spacecraft on the surface of the Moon. The first human-made object to touch the Moon was the Soviet Union's Luna 2, on 13 September 1959. Apollo 11 (July 16–24, 1969) was the American spaceflight that ......\\
\underline{Thought}:\\
\textbf{Output}:\\
Retrieve for more information about Apollo 11\\
\\
\textbf{Plan-Agent (generate Action name)}:\\
\textbf{Input}:\\
(format requirements)
(tool usage instructions)\\
\underline{Question}: The first human-made object to land on the moon, in 1969, was which spacecraft?\\
\underline{Thought}: I should first search the Moon landing history.\\
\underline{Action}: BingSearch[moon landing spacecraft]\\
\underline{Observation}: A Moon landing or lunar landing is the arrival of a spacecraft on the surface of the Moon. The first human-made object to touch the Moon was the Soviet Union's Luna 2, on 13 September 1959. Apollo 11 (July 16–24, 1969) was the American spaceflight that ......\\
\underline{Thought}: Retrieve for more information about Apollo 11.\\
\underline{Action}:\\
\textbf{Output}:\\
Retrieve\\
\\
\textbf{Tool-Agent}:\\
\textbf{Input}:\\
(format requirements)
(tool usage instructions)\\
\underline{Question}: The first human-made object to land on the moon, in 1969, was which spacecraft?\\
\underline{Thought}: I should first search the Moon landing history.\\
\underline{Action}: BingSearch[moon landing spacecraft]\\
\underline{Observation}: A Moon landing or lunar landing is the arrival of a spacecraft on the surface of the Moon. The first human-made object to touch the Moon was the Soviet Union's Luna 2, on 13 September 1959. Apollo 11 (July 16–24, 1969) was the American spaceflight that ......\\
\underline{Thought}: Retrieve for more information about Apollo 11.\\
\underline{Action}: Retrieve\\
\underline{Output}:\\
Apollo 11\\
\\
\textbf{Reflect-Agent (generate Thought)}:\\
\textbf{Input}:\\
(format requirements)
(tool usage instructions)\\
\underline{Question}: The first human-made object to land on the moon, in 1969, was which spacecraft?\\
\underline{Thought}: I should first search the Moon landing history.\\
\underline{Action}: BingSearch[moon landing spacecraft]\\
\underline{Observation}: A Moon landing or lunar landing is the arrival of a spacecraft on the surface of the Moon. The first human-made object to touch the Moon was the Soviet Union's Luna 2, on 13 September 1959. Apollo 11 (July 16–24, 1969) was the American spaceflight that ......\\
\underline{Thought}: Retrieve for more information about Apollo 11.\\
\underline{Action}: Retrieve[Apollo 11]\\
......\\
\underline{Action}: Finish[Apollo 11]\\
\underline{Observation}: Please reflect your answer based on the history.\\
\underline{Thought}:\\
\textbf{Output}:\\
The question asks about the first human-made object to land on the moon, so it seems that the Soviet Union's Luna 2 is more like the answer.\\
\\
\textbf{Reflect-Agent (generate Action)}:\\
\textbf{Input}:\\
(format requirements)
(tool usage instructions)\\
\underline{Question}: The first human-made object to land on the moon, in 1969, was which spacecraft?\\
\underline{Thought}: I should first search the Moon landing history.\\
\underline{Action}: BingSearch[moon landing spacecraft]\\
\underline{Observation}: A Moon landing or lunar landing is the arrival of a spacecraft on the surface of the Moon. The first human-made object to touch the Moon was the Soviet Union's Luna 2, on 13 September 1959. Apollo 11 (July 16–24, 1969) was the American spaceflight that ......\\
\underline{Thought}: Retrieve for more information about Apollo 11.\\
\underline{Action}: Retrieve[Apollo 11]\\
......\\
\underline{Action}: Finish[Apollo 11]\\
\underline{Observation}: Please reflect your answer based on the history.\\
\underline{Thought}: The question asks about the first human-made object to land on the moon, so it seems that the Soviet Union's Luna 2 is more like the answer.\\
\underline{Action}:\\
\textbf{Output}:\\
Reflect[wrong]

\end{document}